\documentclass[lettersize,journal]{IEEEtran}
\usepackage{amsmath,amsfonts}
\usepackage{algorithmic}
\usepackage{array}
\usepackage[caption=false,font=normalsize,labelfont=sf,textfont=sf]{subfig}
\usepackage{textcomp}
\usepackage{stfloats}
\usepackage{url}
\usepackage{verbatim}
\usepackage{graphicx}
\usepackage{cite}
\usepackage{amsmath}
\usepackage{amssymb}
\usepackage{amsthm}
\usepackage{upgreek}
\usepackage{bm}
\usepackage[ruled,vlined,linesnumbered]{algorithm2e}
\usepackage{caption}
\usepackage{subcaption}
\usepackage{enumitem}
\usepackage{multirow,booktabs}
\usepackage{tabularray}
\usepackage{color}
\setlist[itemize]{leftmargin=*}

\DeclareMathOperator*{\argmax}{argmax}
\DeclareMathOperator*{\locmax}{locmax}
\newcommand{\pushline}{\Indp}
\newcommand{\popline}{\Indm}

\let\oldnl\nl
\newcommand{\nonl}{\renewcommand{\nl}{\let\nl\oldnl}}

\SetArgSty{textnormal}  

\SetKw{KwRet}{return}
\SetKw{KwAll}{all}
\SetKw{KwNot}{not}

\SetKwInOut{Input}{Input}
\SetKwInOut{Output}{Output}
\SetKwInOut{Data}{Data}
\SetKwInOut{Result}{Result}
\SetKwInOut{Stop}{Stop}
\SetKwFunction{fLHS}{latinHypercubeSample}
\SetKwProg{fn}{Function}{:}{\KwRet}

\newtheorem{theorem}{Theorem}

\theoremstyle{definition}
\newtheorem{definition}{Definition}

\theoremstyle{remark}

\newcommand{\+}[1]{\ensuremath{\mathbf{#1}}}

\begin{document}

\title{ADs: \underline Active \underline Data-\underline sharing for Data Quality Assurance in Advanced Manufacturing Systems}

\author{Yue Zhao, Yuxuan Li, Chenang Liu, Yinan Wang
\IEEEcompsocitemizethanks{
\IEEEcompsocthanksitem Yue Zhao and Yinan Wang are with the Department of Industrial and Systems Engineering, Rensselaer Polytechnic Institute, Troy, NY, 12180. \protect\\
E-mail: \{zhaoy23, wangy88\}@rpi.edu
\IEEEcompsocthanksitem Yuxuan Li and Chenang Liu are with the School of Industrial Engineering and Management, Oklahoma State University, Stillwater, OK, 74078.\protect\\
E-mail: \{yuxuan.li, chenang.liu\}@okstate.edu}
\thanks{Manuscript received XXX; revised YYY. (Corresponding Author: Yinan Wang)}

\markboth{Manuscript}}


\maketitle

\begin{abstract}
Machine learning (ML) methods are widely used in manufacturing applications, which usually require a large amount of training data. However, data collection needs extensive costs and time investments in the manufacturing system, and data scarcity commonly exists. With the development of the industrial internet of things (IIoT), data-sharing is widely enabled among multiple machines with similar functionality to augment the dataset for building ML models. Despite the machines being designed similarly, the distribution mismatch inevitably exists in their data due to different working conditions, process parameters, measurement noise, etc. However, the effective application of ML methods is built upon the assumption that the training and testing data are sampled from the same distribution. Thus, an intelligent data-sharing framework is needed to ensure the quality of the shared data such that only beneficial information is shared to improve the performance of ML methods. In this work, we propose an Active Data-sharing (ADs) framework to ensure the quality of the shared data among multiple machines. It is designed as a self-supervised learning framework by integrating the architecture of contrastive learning (CL) and active learning (AL). A novel acquisition function is developed for active learning by integrating the information measure for benefiting downstream tasks and the similarity score for data quality assurance. To validate the effectiveness of the proposed ADs framework, we collected real-world \textit{in-situ} monitoring data from three 3D printers, two of which share identical specifications, while the other one is different. The results demonstrated that our ADs framework could intelligently share monitoring data between identical machines while eliminating the data points from the different machines when training ML methods. With a high-quality augmented dataset generated by our proposed framework, the ML methods can achieve a better performance of accuracy 95.78\% when utilizing 26\% labeled data. This represents an improvement of 1.41\% compared with benchmark methods which used 100\% labeled data.

\end{abstract}

\def\abstractname{Note to Practitioners}
\begin{abstract}

This paper is motivated by the need to share data across machines or processes for building machine learning models and the widely existing low-quality data issue. Low-quality data here refers to data samples collected from machines/processes different from the target one. In the manufacturing system, when building a machine learning model for a target machine/system, those low-quality data will decay the model performance if they are selected and shared for training. Therefore, it hinders the direct application of active learning in data sharing as the classic setting of active learning does not consider the impact of data quality on model performance. The objective of this paper is to develop an intelligent data-sharing framework. It is designed to simultaneously select the most informative data points benefiting the downstream tasks and mitigate the impact of low-quality data. We collected real-world \textit{in-situ} monitoring data of the same additive manufacturing process from three different machines, two of which are more similar than the rest. The proposed method is applied to train an anomaly detection model for those two similar machines, and the entire data pool from all three machines is available for selecting and annotating. The results demonstrated that our proposed method outperforms the benchmark methods by only requiring 26\% of labeled training samples. In addition, all selected data samples are from machines with similar conditions, while the data from the different machines are prevented from misleading the training. 
\end{abstract}


\section{Introduction}

Supervised learning has been extensively applied to a wide spectrum of downstream tasks in advanced manufacturing systems \cite{Arinez2020ArtificialII}. For example, 
a machine learning (ML) model might be used to detect manufacturing faults or identify anomalous sensor readings\cite{Abdallah2022AnomalyDA}. The training process of supervised learning requires large quantities of annotated data to achieve good performance on the required task. Unfortunately, in manufacturing systems, it is often impossible to collect a large amount of annotated data due to the high costs of labor, time, and investment, which causes the lack of overall data samples and annotated samples for supervised learning. Therefore, data-sharing is proposed as a solution to share data among multiple manufacturing processes with similar functionality, which naturally augments the amount of data available for the ML model  \cite{shi2023knowledge}. 

Despite the benefits of data-sharing, there are two gaps when applying it in practice. First, although the shared data increases the size of the data pool, they do not necessarily include the most informative subset of data that benefits the model performance on the downstream task, given a limited budget for data annotation. Second, an important assumption to ensure the effectiveness of the ML method is that the training and testing data should be from the same distribution \cite{xu2018splitting}. However, distribution mismatch naturally exists among the collected data from different manufacturing processes or even the same process but slightly different machines. Although with similar functionalities, different processes contain both systematic and stochastic differences in working conditions, process parameters, sensors, machine specifications, etc. Therefore, the objective of data-sharing in advanced manufacturing can be summarized as selecting a subset of data samples across multiple manufacturing processes or different machines sharing identical underly distribution to benefit the performance of downstream tasks under a given annotation budget. In this work, anomaly detection is selected as the downstream task, which is defined to identify normal or abnormal working conditions from multivariate time-series \textit{in-situ} sensor data collected from different manufacturing processes. In the following description, we use the term \textit{target distribution} to refer to the data distribution that the ML methods are designed to model.

The concept of active learning (AL) emerged as a plausible solution to the first gap by prioritizing important data to cut down on annotation costs. Generally, AL identifies a much smaller portion of the data pool for annotation (refer to query set) that maximizes the information gain of the ML model. The idea is that training the model on the annotated subset of data achieves a comparable performance to a model trained on the full dataset. This is usually conducted by quantifying the information measure of the unlabeled data for the ML model. AL techniques have generally achieved great success in these limited-budget tasks and, in some cases, even improved the performance of models with a much smaller amount of training data \cite{Gao_Saar-Tsechansky_2020}. Despite their effectiveness, most of the existing AL methods also assume all the data points are sampled from the same underlying distribution, which does not always hold in practice, especially in the manufacturing system. This is problematic because although the most informative samples would still be selected by the AL method, there is no guarantee that these samples are representative of the target distribution, thus hindering the performance of supervised learning trained on the selected samples. Therefore, it is essential to consider the issue of distribution mismatch when applying AL methods to applications in the manufacturing system. 

\begin{figure}[!tb]
    \centering
    \includegraphics[width=3.475in]{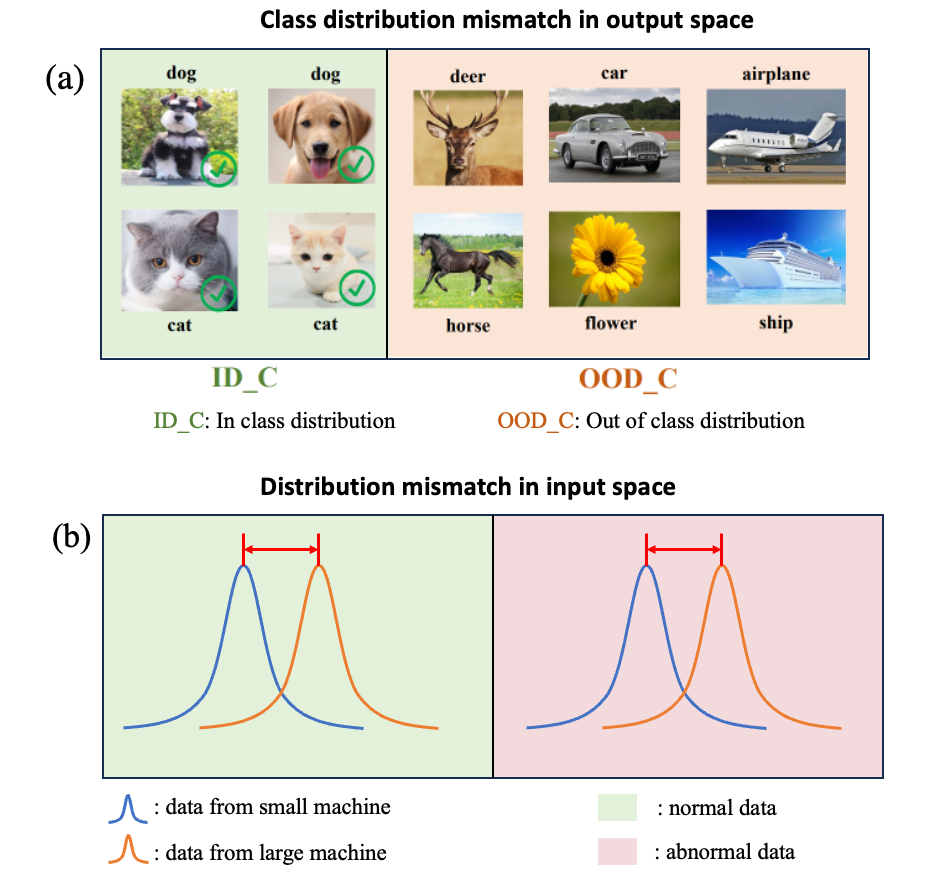}
    \caption{(a) An instance of different class distribution mismatch in output space \cite{du2021contrastive}. (b) An instance of distribution mismatch in input space.}
    \label{fig-inputvsoutput}
\end{figure}

The distribution mismatch over training and testing data has also been observed in various machine learning applications. Du et al. \cite{du2021contrastive} note several typical ML scenarios, i.e., medical diagnoses containing unseen lesions and the house annotation of remote sensing images containing numerous natural sceneries, having the issue of distribution mismatch mitigating the model performance \cite{du2021contrastive}, including classification, regression \cite{pathak2022new}, etc. Applying ML in these situations is referred to as learning under class distribution mismatch, which is naturally a challenging task \cite{yang2015maximum,chen2020semi,guo2020safe}. Therefore, various methods have been developed to resolve the distribution mismatch issue from different perspectives. One popular line of research attacks this issue by correctly detecting the data samples not belonging to the target distribution, which is formally formulated as out-of-distribution (OOD) detection \cite{liang2018enhancing, liu2020energy, wang2023wood}. The idea is to passively eliminate the impact of OOD data by filtering them out in the testing phase. The current OOD detection methods are usually built upon two assumptions: (1) the distribution mismatch mainly existed in the label space, which usually refers to a novel class of object (not existing in the training dataset) appearing in the testing phase for the classification task. This phenomenon is demonstrated in Figure \ref{fig-inputvsoutput}(a), in which novel classes in the cream background might appear in the testing phases; (2) the ``clean" dataset following the target distribution is available when training the ML model. Therefore, in the classification task, the problem of OOD detection is usually formulated as identifying the data samples that do not belong to any class in the training. However, these two assumptions do not hold when sharing data among manufacturing processes due to (1) the distribution mismatch is more likely to exist in the input space due to the systematic and stochastic differences among manufacturing processes, which is demonstrated by the gap between input distributions in blue and orange curves shown in Figure \ref{fig-inputvsoutput}(b); and (2) the ``clean" dataset following the target distribution is usually unavailable for training due to the data scarcity of each single manufacturing process. Consequently, a new paradigm is needed to resolve the challenge of distribution mismatch in data-sharing during model training. 

With emerging attention to mitigate the issue of distribution mismatch during the training process, researchers started to tackle this challenge in the cycle of the AL method. The idea is to actively select the subset of data samples from a general data pool to ensure they both follow the same target distribution and benefit the downstream task. Failure-averse AL is first proposed to incorporate the physical principles into preparing the data samples for the regression task in the manufacturing system \cite{lee2022failure}. Chattopadhyay et al. proposed a novel criterion that achieves good generalization performance of a classifier by selecting a set of query samples to minimize the difference in distribution between the labeled and the unlabeled data \cite{Chattopadhyay}. An integrated information measure is proposed to score and rank the unlabeled data points such that the top candidates are ensured to both benefit the downstream regression task and follow the same physical principle. It provides a way to intuitively incorporate the measure of target distribution into the information measure in AL. However, it still cannot be intuitively applied in a general data-sharing scenario, as the target population usually has no closed-form explicit expression that can be directly exploited. With the recent development of self-supervised learning, contrastive learning (CL) offers the solution to evaluate the similarity over features extracted from different input data \cite{du2021contrastive}. Therefore, dissimilar features naturally correspond to data samples from different input distributions. As a self-supervised learning method, CL does not require label information, which is trained by forming data samples into positive pairs (anchor-positive, i.e., following similar distributions) and negative pairs (anchor-negative, i.e., following dissimilar distributions). The intuitive idea of CL is to train the feature extractor by encouraging it to drag positive pairs close to each other in the feature space while pushing negative pairs away from each other. In that sense, a model trained within the CL framework is capable of embedding data samples with similar distributions close to each other in the feature space, while those with dissimilar distributions are embedded farther away. Thus, it models the distributional differences present in the data. It is worth noting that CL still requires some data samples from the target distribution to initiate the training but is much less data-demanding compared with supervised learning methods due to the advantage of self-supervised schema \cite{möllenbrok2023active}.

Du et al. \cite{du2021contrastive} first propose to incorporate the CL into the cycle of AL to mitigate the distribution mismatch issue in the output space shown in Figure \ref{fig-inputvsoutput}(a), which is referred to as class distribution mismatch in the classification problem. It is not intuitively applicable to our scenario. Similar to our previous discussion, the distribution mismatch in data-sharing mainly exists in the input space. Using anomaly detection as an example, the output space across different manufacturing processes should uniformly contain two classes of normal or abnormal while the monitoring data representing the same working condition (i.e., either normal or abnormal) might follow different distributions across different manufacturing processes due to distribution mismatch in the input space.


To fill in the research gap, the proposed framework, termed Active Data-sharing (ADs), views the problem as a multi-objective AL where the first objective is to select highly uncertain samples for the anomaly detection task and the second is to ensure that the selected data matches the target distribution. The approach is based primarily on the observation that AL and CL can achieve these objectives independently. Therefore, combining them might result in a feasible joint solution. That is, in the context of multi-objective optimization (MOO), the resulting solution would be Pareto optimal. In ADs, each objective is optimized by a separate model, wherein the first involves uncertainty sampling in the form of entropy of classifier predictions, and the second utilizes a CL network that learns similarity features within the data to distinguish data from similar and dissimilar machines. These models are trained on a small set of annotated data as initialization and then applied to the unlabeled data to form two sets of scores corresponding to each objective. A joint query strategy is then used to integrate the two scores into a joint set of scores, which is used to select the best samples for the human annotator. Thus, the proposed framework allows high-quality data-sharing that satisfies both objectives.

The major contributions of the work are: (1) A novel Active Data-sharing (ADs) framework is proposed to ensure the quality of industrial data-sharing when subject to data scarcity, distribution mismatch, and low annotation budget. (2) A novel acquisition function is developed for AL under distribution mismatch in the input space by integrating the informativeness and distribution similarity scores. (3) The effectiveness of the framework is evaluated on real-world \textit{in-situ} monitoring data from additive manufacturing (AM) processes.

The paper is structured in the following manner. Section \ref{section-lit-review} reviews the literature related to AL, CL, and AL under distribution mismatch, which refers to applications where AL in isolation fails to produce desirable results. Section \ref{section-methodology} elaborates on the proposed ADs framework for effective annotation and data-sharing under distribution mismatch. Section \ref{subsection-pareto} presents a formal theoretical background and explanation for the mathematical validity of ADs. Section \ref{section-experiments} evaluates the proposed method in various settings with a comprehensive case study involving a real-world industrial additive manufacturing process by using in-situ monitoring data recorded from three AM processes.

\section{Literature Review} \label{section-lit-review}

This section summarizes the literature and recent work on the key domains of AL, CL, and particularly semi-supervised learning, all of which were critical to developing ADs.

\subsection{Active Learning (AL)}

AL offers various strategies for reducing the annotation budget by actively selecting the most informative or valuable data points and feeding them to the annotator for label acquisition \cite{settles2009active}. These strategies fall into three categories: query synthesis (conventional \cite{king2004functional, wang2015active, schumann2019active}, generative \cite{zhu2017generative, liu2019generative, tran2019bayesian}), sequential sampling \cite{cohn1994improving, dasgupta2007general}, and pool-based sampling \cite{cohn1996active, schein2007active}.

The problem presented in this work belongs to pool-based sampling wherein the unlabeled data pool is the collected \textit{in-situ} monitoring data from three AM processes, and the goal is to evaluate the information measure over the entire unlabeled data pool prior to querying the candidate samples for sharing. Almost all pool-based methods score the samples based on their \textit{informativeness}. The key idea of these approaches is to select only the most informative subset of samples so as to maximize the information gain for the model. Examples include uncertainty sampling approaches \cite{lewis1994heterogeneous, lewis1995sequential, joshi2009multi, luo2013latent, wang2016cost, sinha2019variational, yoo2019learning}, variance reduction \cite{cohn1996active, schein2007active}, query-by-committee \cite{seung1992query, mccallum1998employing, burbidge2007active}, etc. A sample's uncertainty may be quantified in several ways, e.g., marginal uncertainty \cite{roth2006margin} or, most popularly, entropy \cite{shannon1948mathematical}, which utilizes the posterior probability of the model's prediction on all samples to select the best one. This is particularly intuitive if the data can be represented as a probabilistic distribution. Recently, Sinha et al. \cite{sinha2019variational} employed variational autoencoders to determine uncertainty by comparing the distribution of the annotated and unlabeled data. A task-agnostic approach by Yoo et al. \cite{yoo2019learning} proposes to use a parametric loss predictor module to predict the loss of the unlabeled sample and, therefore, measure its uncertainty.

Another class of pool-based AL methods is based on \textit{representative} measures, such as diversity \cite{yang2015multi, sener2017active, kim2022defense}, density \cite{dasgupta2008hierarchical, zhu2009active, wang2017active} or a combination of the two \cite{xu2007incorporating}. Diversity methods favor exploration and prefer the selection of dissimilar instances, whereas density-based approaches assume that either dense or sparse groups of data points contain the most information. Therefore, they prefer selecting instances that are either similar or dissimilar to several other instances \cite{pereira2019empirical}. Hierarchical clustering \cite{dasgupta2008hierarchical} and density estimation methods \cite{zhu2009active, wang2017active} are commonly used for density-based approaches. Amongst diversity-based approaches, the classical core-set approach \cite{wei2013using} is the most popular -- it aims to identify a diverse set of samples, i.e., the core or cover set, by minimizing the distance between each sampled point and the remaining points. The expected result is that a model trained on the core-set is at least equivalent to a model trained on the complete data in terms of performance. The core-set was first adapted to batched inputs for convolutional neural networks (CNNs) by Sener et al. \cite{sener2017active}. They proposed a robust k-center algorithm operating on Euclidean distances of the last layer's feature vectors for a batch of input images. Kim et al. \cite{kim2022defense} propose a density-aware core-set approach (DACS) to select diverse samples from locally sparse regions, which is useful if the sparse regions contain informative samples compared to densely grouped instances.  

An inherent flaw of the representative approaches is that all samples are treated equally in terms of informativeness, which is not necessarily the case. It is also possible, however, that a sample's representativeness is related to its importance, in which case the representative measures might be more effective. Recently, it has been verified that a combination of both informative and representative approaches results in better performance \cite{zhu2009active, huang2010active, huang2013active, wang2015querying, lin2017active, tang2019self}.


However, the underlying assumption of the active-learning-based methods, as well as most other methods focusing on maximizing information gain, is that the distribution of the data is the same, whether the samples are annotated or not. Upon violation of this assumption, the performance of these methods deteriorates sharply \cite{du2021contrastive}. Recall that class distribution mismatch and mitigate it in data-sharing tasks is a central focus of this work. Therefore, even though pool-based uncertainty sampling seems to be a viable solution for at least picking informative data points, it is not advisable to apply it unless the mismatched data is somehow identified and excluded.

\subsection{Contrastive Learning (CL)}

CL is a self-supervised, task-agnostic technique to learn effective high-dimensional feature representations of data, usually to the benefit of a downstream task. Practically, CL is utilized to improve the performance of a model by exposing it to pairs of annotated negative and positive samples in addition to actual samples (anchors) to produce high-dimensional embeddings of the data and designing the loss function \cite{hadsell2006dimensionality, schroff2015facenet} so as to maximize positive-anchor and minimize negative-anchor distances. The labels \textit{positive} and \textit{negative} are termed the similarity labels in this context. In practice, CL has been successfully applied to various tasks related to vision, natural language processing, etc. It greatly improves model performance under distribution mismatch or when the inputs share similar features but belong to distinct classes \cite{jaiswal2020survey}.

Owing to its widespread success in self-supervised learning and task-independent nature, the technique has been applied to several domains and continues to receive significant attention and development. Transformations of the input data to generate positive augmentations of anchors are crucial, and the type of transformation(s) used should be based on the format of the data. For example, SimCLR \cite{chen2020simple} composes several different types of augmentations via local and global image transformations such as crops, rotations, cutouts, blurs, color distortions, etc., and additionally provides an account of augmentations that, when considered positive, seem to worsen performance. Building off of these findings, CSI \cite{tack2020csi} proposed to term these problematic augmentations as distribution-shifting transformations and instead feed them as negative samples to the CL framework. This led to further study on learning shift-invariant feature representations by integrating CL and clustering approaches \cite{caron2020unsupervised, li2020prototypical}. The fully unsupervised Winner-Take-All (WTK) autoencoder architecture \cite{makhzani2015winner} enforces sparsity in addition to shift-invariance of the learned representations, which can be used to reliably generate in-distribution augmentations. Furthermore, the WTK architecture allows joint back-propagation of gradients through both the encoder and decoder paths, which results in quicker and more stable training.

It is worth noting that in the absence of similarity labels, CL becomes susceptible to \textit{sampling bias}, which leads to the inclusion of false negatives because, in the simplest case, negative samples for an anchor are randomly picked from the dataset. While similarity labels are assumed available for a subset of the data in ADs, Chuang et al. \cite{chuang2020debiased}, and several others \cite{zhao2021graph, zhou2022debiased} propose debiased CL frameworks for specific downstream tasks that focus on minimizing sampling bias for fully unsupervised CL.

\subsection{Constrained Active Learning}

AL approaches generally have a drawback that they are inherently unaware of the underlying distribution of the data and, therefore, cannot be used when the data suffers from class distribution mismatch. At the same time, applying CL approaches to the described problem is not a plausible solution either because (1) being task-agnostic, they do not extract features specifically for improving the performance on the downstream task, and (2) human annotation can resolve the issue of lacking labels but cannot identifying distribution mismatch, while the CL can identify distribution mismatch. 


Several papers in recent years have proposed and developed constrained AL techniques in order to leverage the benefits of AL in certain feasible regions. Constrained AL can mitigate the decay on downstream task performance when directly applying AL \cite{du2021contrastive, margatina2021active, lee2022failure}. Du et al. \cite{du2021contrastive} propose CCAL, an AL framework with a joint query strategy wherein each sample is scored on the basis of two separate scores that are then combined into a joint score used to select the best samples. They also prove a tight upper boundary on the consequent error function termed the CCAL error and compare the results for the task of image classification under distribution mismatch with the two other semi-supervised learning methods (DS\textsuperscript{3}L \cite{guo2020safe} and UASD \cite{chen2020semi}) under distribution mismatch. It should be noted that semi-supervised learning is different from AL in that it directly utilizes the unlabeled data during training. In addition, no extra annotation efforts will be applied to the unlabeled data.

Lee et al. \cite{lee2022failure} explored incorporating physical constraints into AL to accommodate a more practical context, such as an engineering system, where physical constraints are present. Violation of these constraints could result in fatal system failure. The authors suggest the existence of a safe region in the sample space and attempt to make AL focus on exploring the safe region. At the same time, it is desirable to explore the safe region as thoroughly as possible with limited data samples. The authors propose the PhyCAL framework, utilizing safe variance reduction and safe region expansion to trade-off between information maximization and safe exploration of the design space.

The ADs framework proposed in this paper expands on these methods. More specifically, (1) instead of AL for image classification under distribution mismatch over the output space, ADs focus on resolving the distribution mismatch in the input space when sharing data over multiple manufacturing processes, (2) a new joint query strategy is proposed to select the samples simultaneously following the target distribution and benefiting the downstream tasks. In addition, the convergence of the joint function to the optimal point is proven with methods pertaining to convex analysis and Pareto optimal.

\section{Methodology} \label{section-methodology}

\begin{figure}[!tb]
    \centering
    \includegraphics[width=3.4in]{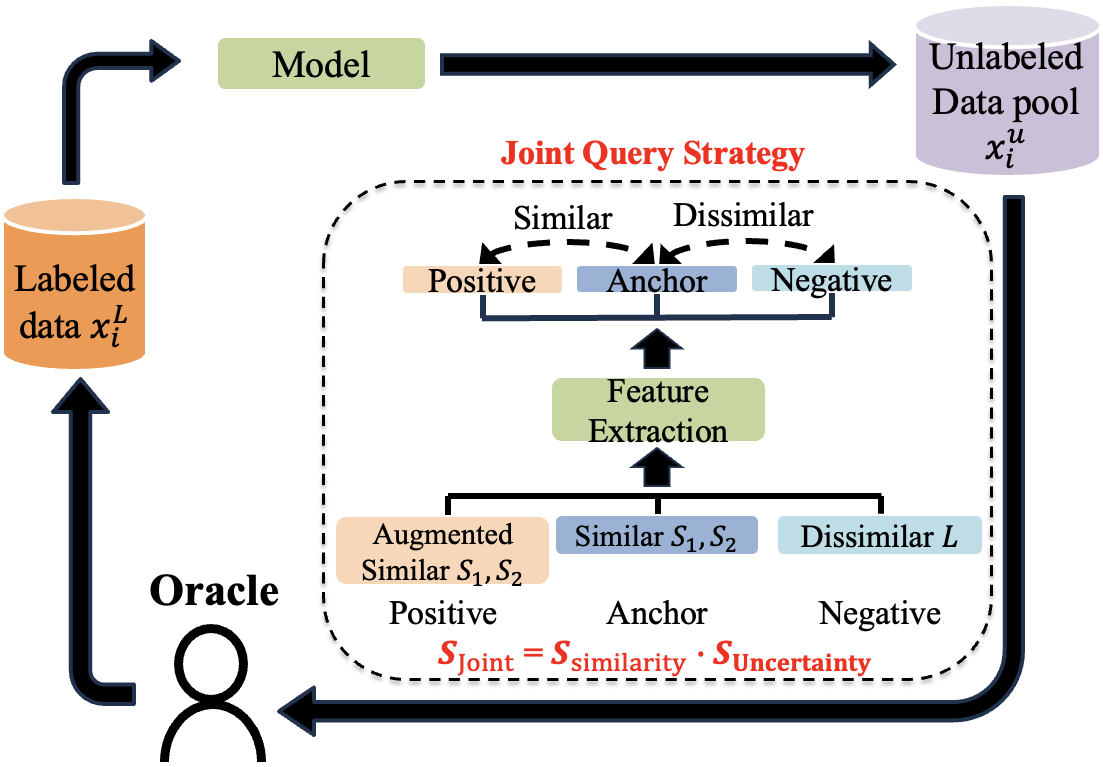}
    \caption{Architecture of active data sharing (ADs).}
    \label{fig-active-learning-loop}
\end{figure}

This section comprehensively explores the proposed ADs framework to solve the problem of data sharing among multiple manufacturing processes to jointly select the most informative samples for the downstream task and mitigate distribution mismatch over the input space. First, the general notations and nomenclature are introduced. Then, the calculation of the individual scores and their integration into the joint score is explained. Finally, a theoretical investigation of the Pareto optimality of the proposed method is conducted through convex analysis of the loss functions with the proposed individual and joint acquisition functions. For brevity, the data pre-processing stage and data augmentation model are left for Section \ref{section-experiments}, where data management is more relevant.

This study involves data-sharing on \textit{in-situ} monitoring data among various manufacturing processes assigned to produce identical objects. Here, we assume small and large machines are used respectively and are the main reason incurring the difference among manufacturing processes. Monitoring data collected from small machines with identical manufacturers and model numbers are under similar distributions, while those from large machines with different manufacturers and model numbers are recognized as having dissimilar distributions. Consequently, a distribution mismatch is observed in the input space. The downstream task of the study is anomaly detection, which distinguishes between normal and abnormal working conditions of the manufacturing process.

Consider that machines $S_1$ and $S_2$ are similar, providing data that follows a similar distribution. Suppose the data from the dissimilar machine $L_1$, with a mismatched distribution relative to $S_1$ and $S_2$, is additionally introduced to form a combined dataset $\mathcal{D} = (S_1,S_2,L_1)$. Let $\mathcal{L} = \{\text{normal},\,\text{abnormal}\}$ be the labels or classes of each sample associated with the downstream task, regardless of which machine they were sourced from. Let $\+{X} \in \mathbb{R}^{D\times 3}$ be the matrix representation storing all samples in $\mathcal{D}$. Then, $\+X^{S_1}, \+X^{S_2}$ and $\+X^{L_1}$ are all disjoint sub-matrices representing the data from each machine. As they are disjoint, the concatenation and disjointed union (or logical XOR) of any combination of these subsets is equivalent, i.e., $\+X^{S_1} \frown \+X^{S_2} = \+X^{S_1} \oplus \+X^{S_2}$, etc. Finally, suppose that a human annotator labels a small portion of these samples as normal and abnormal. Thus, there is a small annotated pool of samples in the form of a matrix $\+D_{\text{label}}$, and a large pool of unlabeled data $\+D_{\text{unlabel}} = \+X \setminus \+D_{\text{label}}$, such that $\+D_{\text{unlabel}} >> \+D_{\text{label}}$, i.e., the unlabeled portion is much larger than the annotated portion. The purpose of ADs is to extend AL to form a query set $\+D^Q$ from the unlabeled data pool $\+D_{\text{{unlabel}}}$, such that each queried sample, $\+x_i^Q$ is not only among the most informative samples in $\+D_{\text{unlabel}}$, but also has a high likelihood of belonging to the similar machine data:
\[ \forall \; \+x_i^Q \in \+D^Q: \underbrace{\text{entropy}(\+x_i^Q) \to 1}_\text{high uncertainty} \;\wedge\; \underbrace{\+x_i^Q \in (\+X^{S_1} \oplus \+X^{S_2})}_\text{high similarity} \]

In order to satisfy both objectives, the joint query strategy employed in ADs consists of integrating two separately calculated scores, termed the similarity and uncertainty scores, to form a joint acquisition function $\+J$, which scores the unlabeled data based on how well they satisfy both objectives simultaneously. The individual scores are computed based on the feature representations obtained by two separate models. Specifically, the similarity score is obtained via the feature vector extracted from the CL model, and the uncertainty score is obtained via the current anomaly detection model. 

For the subsequent subsections, let feature $Z_s(\cdot)$ and feature $Z_u(\cdot) \in [0,1]$ denote a forward pass through the CL and uncertainty sampling networks to generate a feature vector, and let $D_l$ and $D_u$ be the number of samples in the labeled and unlabeled data pools respectively.

\subsection{Learning Similarity Features} \label{subsection-contrastive-model}

The CL model is trained to quantitatively evaluate the similarity between data from the target distribution ($S_1 \cup S_2$) and data from a different distribution ($L_1$). In this sense, it can be further exploited to facilitate the objective that queried samples $\+D^Q$ from the unlabeled pool $\+D_{\text{unlabel}}$ closely match the target distribution of data $X^{S_1} \oplus X^{S_2}$. In our application, it indicates that for each cycle of AL, all the selected samples would be from similar machines. This naturally leverages the fact that the distribution of \textit{in-situ} monitoring data varies slightly between the similar agents $(S_1,S_2)$, but considerably in the case of $(S_1,L_1)$ or $(S_2,L_1)$.

To prepare positive and negative pairs of samples for training purposes, the initially sampled and annotated data from $S_1 \cup S_2$ are fed into a data augmentation model (trained as part of the preliminary processing detailed in Section \ref{subsection-wtk}) to generate two positive augmentations based on the augmentation Equations (\ref{augmentation-1}), (\ref{augmentation-2}). These augmentations are used as the positive samples while the original data sample is kept as the anchor. Similarly, the initially annotated samples from L1 are randomly chosen as negative samples.

With the prepared positive and negative pairs of samples, triplet loss function \cite{schroff2015facenet} is used to train the CL model to better differentiate samples belonging to different distributions. Its input consists of positive and negative samples for the same anchor, which is the data sample from the target distribution. Additionally, a distance function must be defined to quantify the similarity between positive (anchor versus positive sample) and negative (anchor versus negative sample) pairs. Regardless of the choice of distance function, the triplet loss is optimized toward maximizing the similarity for the positive pair and minimizing the similarity for the negative pair. The expression of triplet loss is given as follows 
\begin{equation}
    \label{general-triplet-loss}
    l_{\text{triplet}}(\+x_a, \+x_p, \+x_n) = \max(f_d(\+x_a, \+x_p) - f_d(\+x_a, \+x_n) + \epsilon ,\,0),
\end{equation}

where $\+x_p$ is a positive sample, $\+x_a$ is the anchor, and $\+x_n$ is a negative sample; $f_d$ is the general distance function that can be applied to each pair; $\epsilon$ is a small positive constant added for numerical stability. 
In this work, cosine similarity is used as the distance function. The triplet loss in Equation (\ref{general-triplet-loss}) for sample $\+x_j \in \+D_\text{label}$ where $j = \{a, p, n\}$ is then:

\begin{equation}
    \label{cosine-dist-triplet-loss}
    l_{\text{triplet}}(\+x_a, \+x_p, \+x_n) = \max(\cos(\+x_a, \+x_n) - \cos(\+x_a, \+x_p) + \epsilon,0)
\end{equation}

Assuming a batch of data samples $\mathcal{B} = \{x\}_{i=1}^{B}$, where $B$ is the number of samples in the batch, the loss function in Equation (\ref{cosine-dist-triplet-loss}) is updated as the average loss over all the samples in the batch:
\begin{equation}
    \label{cosine-dist-triplet-loss-batched}
    l(\mathcal{B}) = \frac{1}{B} \sum_{i=1}^{B}l_{\text{triplet}}(\+x_{ai}, \+x_{pi}, \+x_{ni})
\end{equation}

The contrastive model needs to be trained on a small initial subset of annotated data to be capable of evaluating the similarity among incoming sample pairs. For each unlabeled sample $\+x_i^U \in \+D_{\text{unlabel}} \text{ where } i = \{1,\ldots,d_u\}$, the corresponding similarity score $s_i$ is obtained as follows: (1) All annotated samples $\+D_{\text{label}} \in \mathbb{R}^{d_l\times 3}$ are passed through the trained contrastive model, generating an array of feature vectors $\+F^L = Z_s(\+D_\text{label}) \in \mathbb{R}^{d_l\times d}$, where row $j$ denotes a feature vector $\+f_j^L \in \mathbb{R}^{d}$; (2) unlabeled sample $\+x_i^U$ is passed through the trained contrastive model, generating a $d$-dimensional feature vector $\+f_i^U = Z_s(\+x_i^U)$; (3) The pairwise cosine similarity between $\+f_i^U$ and $\+F^L$ is computed to obtain a vector of cosine similarities $\+c$ with $d_l$ elements and an arbitrary element is $\cos(\+f_j^L, \+f_i^U) \in [-1,1]$. (4) Maximum value in $\+c$ is identified as the final similarity score $s_i = \max \+c$ for the incoming sample $\+x_i^U$, which is expressed as 
\begin{equation}
    \label{sim-score-eq}
    s_i = \max_{j=\{1,...,d_l\}} \, \cos\left(Z_s\left(\+x_j^L\right), Z_s\left(\+x_i^U\right)\right) 
\end{equation}

A higher similarity score corresponds to an increased probability of the unlabeled data point belonging to the target distribution. Notice that this score involves both the annotated data pool and the unlabeled data in the forward pass. To get the complete vector, steps 2--4 are repeated for all remaining samples in the unlabeled data pool $\+D_\text{unlabel}$, resulting in a vector of contrastive similarity scores $\+S^\prime$ through iterative concatenation $\+S^\prime := \+S^\prime \frown s_i,\; \forall i$:

 The CL model is applied to evaluate the similarity of semantic features among data points and elevated similarity scores for data originating from a shared target distribution. This enables us to selectively choose datapoints within the similar distribution (smaller machines). This selection is visually represented in Figure \ref{fig-multi-objective-classification} by the mauve region.

\subsection{Uncertainty Sampling}

The entropy for unlabeled samples can be used to compute the informativeness of that sample. Entropy sampling is a popular AL technique \cite{settles2009active} under the category of uncertainty sampling approaches. Samples exhibiting high uncertainty are ideal samples for annotation as they lie near the decision boundary and thus contain potentially valuable information for the model to learn. For a binary classification problem where $x$ is a sample, and $\hat y$ is the predicted class, entropy is defined as:
\begin{equation}
    \begin{aligned}
    H(x) & =  P_\theta(\hat y\;|\;x)\log\left(\frac{1}{P_\theta(\hat y\;|\;x)}\right)&\\
    & = -P_\theta(\hat y\;|\;x)\log(P_\theta(\hat y\;|\;x))
    \end{aligned}
\end{equation}

The computation of the vector of uncertainty scores $\+U \in [0,1]^{d_u}$ is now discussed. Naturally, it involves the uncertainty model described above, trained on the same initial subset of annotated data as the CL model. Consider a sample from the unlabeled data pool $\+x_i^U \text{ where } i = \{1,\ldots,d_u\}$. Let $c_i^U$ represent its predicted class label $\argmax_k Z_u(x_i^U)$ for $k \in \mathcal{L}$. In addition, suppose $p_{i,k}^U$ represents the normalized probability that a sample $x_i^U$ belongs to class $k$, so that $p_{i,k}^U = p({g^U_i} = k \;|\; \+x_i^U)$, where $g^U_i$ represents the probability of any class. The unlabeled sample $\+x_i^U = (x_i, y_i, z_i)$ is passed through the trained classifier model to generate the features $Z_{u}$, and Shannon entropy $H$ is calculated for $\+p_i^U$, which represents the uncertainty score $u_i$. 
\begin{equation}
    \label{uncertainty-score-eq}
    u_i = H \left( Z_u \left( \+x_i^U \right) \right) = H(\+p_i^U) = - \sum_k p_{i,k}^U \log_2 p_{i,k}^U
\end{equation}
The vector of uncertainty scores is also generated iteratively through $\+U := \+U \frown u_i, \; \forall i$. Therefore, Uncertainty sampling aims to select instances that improve model performance by focusing on the most informative or uncertain datapoints. The uncertainty score guides the selection of data points within the blue region depicted in Figure \ref{fig-multi-objective-classification}b. This strategy is particularly useful in scenarios where labeling data is expensive or time-consuming, as it helps make the most out of the limited labeled data available.

\subsection{Joint Query Strategy} \label{subsection-joint-score-comp}

\begin{figure}[!t]
    \centering
    \includegraphics[width=3.4in]{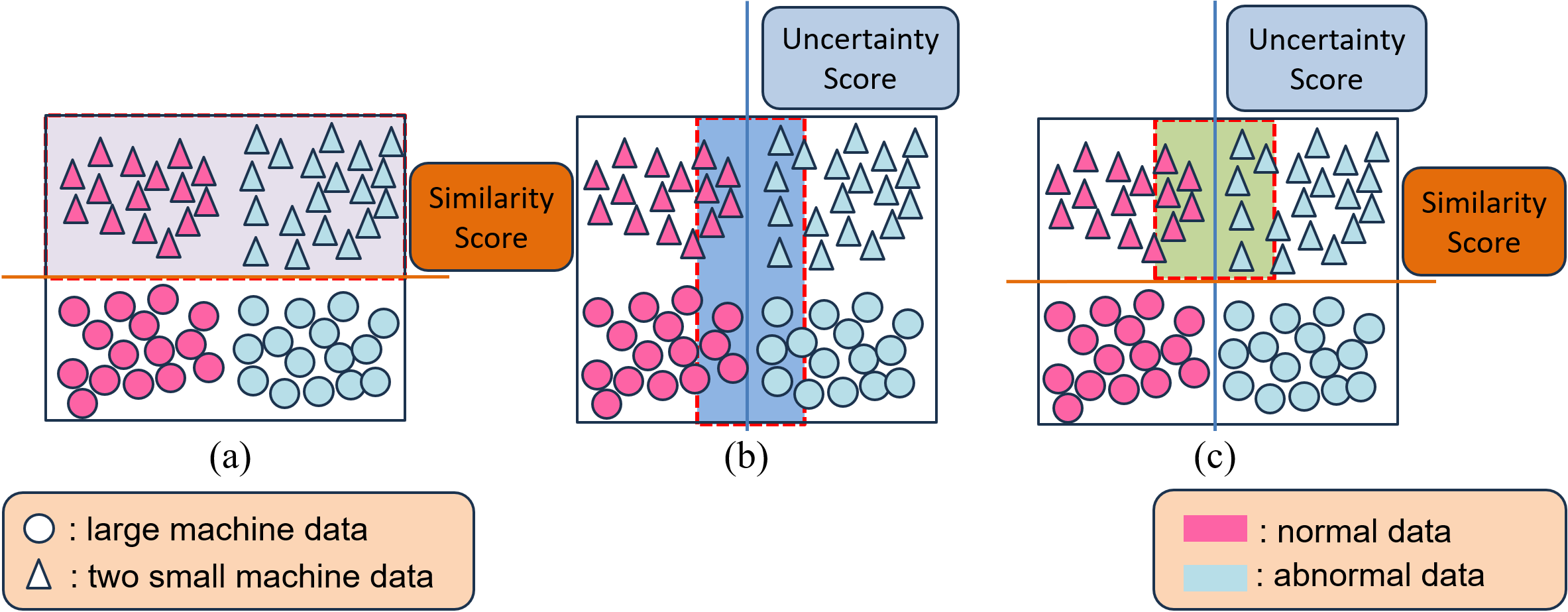}
    \caption{Visualizing divisions of the sample space by applying the ADS framework applied to the unlabeled data pool. (a) The similarity score $\+S$ based on the contrastive model forms a horizontal orange decision boundary in the sample space, addressing the distribution mismatch in the data by separating the mismatched data of the large machines (circles) from the desired data of the small machines data (triangles, mauve region). (b) The uncertainty score $\+U$ based on the entropy of the classifier scores (normal/abnormal) as the blue region containing the desired subset of highly informative samples. The vertical blue decision boundary is generated by the binary classifier. (c) The best joint scores in $\+J = \+S \odot \+U$ identify the samples that are both highly informative for the downstream task and follow the desired similar data distribution from the two small machines (green region).}
    \label{fig-multi-objective-classification}
\end{figure}

In the framework of AL, the idea is to iteratively select and annotate a certain number of informative samples in each cycle to improve the performance of the downstream task. In our problem, the unlabeled samples are evaluated from two aspects: (1) the closeness to the target distribution that is evaluated by the similarity score, and  (2) the potential improvement to the anomaly detection task that is evaluated by the uncertainty score. Guided by the idea that we want to prevent the distribution-mismatch issue in data sharing, the joint query strategy is designed to select the samples with a high uncertainty score (benefit the downstream task) conditioned on they are close to the target distribution (with a high similarity score). 

Therefore, the similarity score $\+S^\prime$ is firstly binarized to set samples with the top $w\%$ similarity scores as 1 and the remaining as 0. The binarized similarity score is denoted as $\+S$. The binarization will be repeated in each cycle of the proposed ADs. Therefore, the value of $w$ is determined to make sure that there are enough samples to be queried. This \textit{binarization} of $\+S^\prime$ is also motivated by the fact in the case study that the number of negative samples $\ell(\+X^{L_1})$ might be larger than the number of positive samples $\ell(\+X^{S_1} \oplus \+X^{S_2})$ in the given dataset. Given this situation, we tend to set the value of $w$ as small as possible to ensure only samples close to the target distribution (with the highest similarity scores) are selected. With these two criteria, the value of $w$ is determined by ensuring there are \textit{just enough} samples with the binarized similarity score of 1 to be queried. This work quotes 25\% of the amount of unlabeled data as the choice of $w$ that consistently satisfied the criterion for the various test settings.

The binarization described above is formally defined as a function $\locmax_n$ that maps an $m$-dimensional vector $\+X$ to another $m$-dimensional mask vector $\+Y$, such that each element of $\+Y \in \{0, 1\}$ denotes whether the corresponding element of $\+X$ was part of the top-$n$ maximum set (i.e., 1) or not (i.e., 0). For example, $y_1 = 1$ indicates the $x_1$ has the top-$n$ maximum in $\+X$, otherwise $y_1 = 0$. Then, for $i = 1,2,\ldots,n$:
\begin{equation}
    \label{eq-locmax}
    \begin{aligned}
    &\locmax_n: \\&\+X \in \mathbb{R}^m \mapsto \+Y \in \mathbb{R}^m \;|\; y_i = 1 \text{ iff } x_i \in \max_n(\+X) \text{ else } y_i = 0
    \end{aligned}
\end{equation}
where $\max_n$ is simply a vector of $n$ maximum values of a target vector. It can be formally defined as:
\begin{equation}
\begin{aligned}
\label{eq-nmax}
    \max_n :{} & \+X \in \mathbb{R}^m \mapsto \+Y \in \mathbb{R}^n \;|\; n < m \\ 
    & \+X_0 = \emptyset \\ 
    & \+X_{n+1} = \+X_n \cup \{\max(\+X \setminus \+X_n)\} \text{ for } n \geq 0
    \end{aligned}
\end{equation}

The binarized similarity score is then given as $\+S = \locmax_{\lfloor k\times d_{u} \rfloor}(\+S^\prime)$. The joint score can now be computed. Observe that there are two $d_u$-dimensional vectors $\+S$ and $\+U$ representing the similarity and uncertainty scores for the unlabeled data pool. The joint score vector $\+J$ is now computed by calculating their Hadamard product:
\begin{align}
    \label{joint-score-vec-eq}
    \+J &= \+S \odot \+U
\\
    \label{joint-score-eq}
    \left(J_1,\ldots,J_{d_u}\right) &= \left(s_1,\ldots,s_{d_u}\right) \odot \left(u_1,\ldots,u_{d_u}\right)
\end{align}
Given that $s_i \in \{0,1\}$ and $u_i \in [0,1]$, it follows that $J_i = s_i u_i \in [0, 1]$. The higher the joint score $J_i$ for a sample, the more likely that it is both similar to the datasets $X^{S_1} \oplus X^{S_2}$, and has high entropy. Thus, by selecting samples with high joint score values, we effectively query samples for annotation that are (1) highly likely to belong to the desired similar machine distribution, and (2) benefit the downstream anomaly detection task, i.e., they lie close to the classification boundary. 
Our objective is to integrate the architectures of CL and AL, capitalizing on the unique strengths of each approach. In CL, we measure the similarity of semantic features among data points, assigning high similarity scores to data from the same underlying distribution. This similarity score is then employed to selectively pick data points from a smaller machine (the mauve region in Figure \ref{fig-multi-objective-classification}a) that shares a similar distribution. In addition, AL utilizes entropy to score the least certain instances for data point selection(the blue region in Figure \ref{fig-multi-objective-classification}b). This uncertainty measure proves valuable for distinguishing between normal and abnormal data during classification. This strategy is particularly useful in scenarios where labeling data is expensive or time-consuming. Combine both the high similarity scores and high uncertainty scores to select samples for labeling (the green region in Figure \ref{fig-multi-objective-classification}c).

\subsection{Main Active Learning Cycle}

\begin{algorithm} 
\caption{Active Data-sharing} \label{algo-ADs}
\DontPrintSemicolon

\Input{Small machine labeled data $D^S_\text{label}$: $\{S_{1}^{X},S_{1}^{Y}, S_{2}^{X},S_{2}^{Y}\} $\\

Large Machine labeled data  $D^L_\text{label}$:$\{L_{1}^{X},L_{1}^{Y}\}$\\

Unlabeled data pool $D_\text{unlabel}$} 

\Output{Trained classifier $\theta_u$}

\BlankLine
\BlankLine

Train the similarity model $\theta_s$ with $\{D^S_\text{label}, D^L_\text{label}\}$ \;
\pushline \nonl
\popline

\For{$I \  in \  AL cycle$}{
    
    Train the classifier Model $\theta_u$ with $D^S_\text{label}$ \;

    Calculate the uncertainty feature of $D_\text{unlabel}$ using $\theta_u$: $z_u({D_\text{unlabel}})= \theta_u(D_\text{unlabel}$) \;
    
    Calculate $S_\text{uncertainty}(D_\text{unlabel})$ using Eq. (\ref{uncertainty-score-eq});   

    Calculate the similarity feature of $D^S_\text{label}$ and $D_\text{{unlabel}}$ using $\theta_s$: $z_s (D^S_\text{{label}})  = \theta_s (D^S_\text{label}) ,z_s (D_\text{unlabel}) = \theta_s (D_\text{unlabel})  $ \;

    Calculate $S_\text{similarity}(D_\text{unlabel})$ using Eq. (\ref{joint-score-vec-eq})\;   
    
    Calculate joint score $J_\text{unlabel}$ using Eq. (\ref{joint-score-eq}) with $S_\text{uncertainty}(D_\text{unlabel})$ and $S_\text{similarity}(D_\text{unlabel})$ \;
    Query the samples with max $J_\text{unlabel}$ to query set $X_\text{Queried}$\;
    Request oracle to annotate all labels in $Y_\text{Queried}$\;
    $D_\text{labeled}^S \gets D_\text{labeled}^S \cup \{X_\text{Queried}, Y_\text{Queried}\}$
    }
\Return{Trained classifier $\theta_u$}
\end{algorithm}

The Algorithm \ref{algo-ADs} for our proposed ADs framework is presented in this subsection and also shown in Figure \ref{fig-active-learning-loop}. Initiated with Latin Hypercube Sampling (LHS), we meticulously select the initial labeled dataset from two small machines $D^S_\text{label}$: $\{S_{1}^{X},S_{1}^{Y}, S_{2}^{X},S_{2}^{Y}\} $ and a large machine $D^L_\text{label}$:$\{L_{1}^{X}, L_{1}^{Y}\}$. This ensures that the set of random numbers represents the genuine variability of the initial dataset. Both pre-trained CL models are trained using this initial labeled data. The steps are summarized as follows:
\begin{enumerate}
    \item In the AL phase, the classifier is trained using the initial labeled data.
    \item Leveraging existing classifiers and pre-trained CL, we extract the similarity feature and uncertainty feature for both labeled and unlabeled data. 
    \item The features are used to calculate the similarity score with Equation (\ref{sim-score-eq}) and uncertainty score with Equation (\ref{uncertainty-score-eq}) of unlabeled data $D_\text{unlabel}$, which are then combined using Equation (\ref{joint-score-vec-eq}). 
    \item The data annotator (oracle) labels the queried data $\{X_\text{Queried}, Y_\text{Queried}\} $ with the highest joint score. 
    \item The labeled dataset undergoes an update by incorporating the queried data, and the size of the annotated data pool increases while the size of the unlabeled data pool decreases by the same amount. 
    \item Signifying the inception of a cyclic process that iteratively repeats steps 1-5. Notably, the classifier model evolves with each cycle, updating dynamically with new labeled data from the oracle.
\end{enumerate}

\subsection{Pareto Optimality of ADs} \label{subsection-pareto} 

It is evident that the described problem can be modeled as a multi-objective optimization problem (MOO) due to the distinct nature of the two objectives involved. Such problems typically have multiple optimal solutions, and it becomes necessary to define the concept of Pareto optimality. 

\begin{definition}[Pareto Optimal \cite{Marler2004SurveyMOO}] 
A point, $x^* \in X$, is Pareto optimal iff there does not exist another point, $x \in X$, such that $F(x) \leq F(x^*)$, and $F_i(x) < F_i(x^*)$ for at least one function $F_i$.
\end{definition}

\begin{definition} [Weakly Pareto Optimal \cite{Marler2004SurveyMOO}]
A point, $x^* \in X$, is weakly Pareto optimal iff there does not exist another point, $x \in X$, such that $F(x) < F(x^*)$.
\end{definition}


For the first objective, i.e., to differentiate data samples from different distributions, the acquisition function is a similarity score $\+S$ based on the cosine similarity metric. It is important to note that this could be any similarity metric, and the approach would still remain valid. To achieve the second objective, i.e., to decide the informativeness of samples, uncertainty sampling \cite{settles2009active} is leveraged to compute the uncertainty score $\+U$. By harmonizing these distinct objective functions to form an integrated criterion that represents both objectives, the queried data $\+D^Q$ meeting the Pareto optimality can ensure meeting both objectives. To that end, the $\+J$ in Equation (\ref{joint-score-vec-eq}) is proposed, forming a scalarized combination of the two individual acquisition functions using element-wise multiplication. In effect, $t$ target samples will be selected for the annotator based on the calculated $\+J$ at the end of each iteration in the active learning. In the following description in this section, we will prove that the Pareto optimal of two objectives in our formulation exists and can be achieved by optimizing the proposed joint acquisition function $\+J$.

\textbf{\textit{Convex Analysis of Joint Acquisition Function}:} The joint acquisition function $\+J$ is defined as the Hadamard product between the similarity score $\+S$ and uncertainty $\+U$. The definition of similarity score (shown in Equation (\ref{sim-score-eq})) indicates that for the $i_{\text{th}}$ sample in the unlabelled data pool, its similarity score ($s_i$) is the maximum of cosine similarities to all the annotated samples. Once $s_i$ has been computed for all samples to compute the vector of cosine similarities $\+S^\prime$, it is further processed by setting $k$\% maximum scores to one and the rest to zero, resulting in a sparse vector $\+S$. Because of this, the similarity score vector is essentially just an indicator vector to select the subset of unlabelled samples that are close to the target population. In addition, the uncertainty score vector $\+U$ (shown in Equation (\ref{uncertainty-score-eq})) is defined upon entropy, which is a concave function (The proof is detailed in Appendix \ref{proof-entropy-concavity}). Additionally, recall that the similarity score vector $\+S$ is a sparse indicator vector so that the Hadamard product representing the joint score function $\+J = \+S \odot \+U$ essentially reduces to the entropy values for selected samples. In other words, the joint score $J_i = s_i u_i$ for a sample is the product of a scalar with a concave function. Therefore, the joint acquisition function is also concave given the following Theorem \ref{theorem-canonical-combinations}.

\begin{theorem}[Canonical Combinations of Convex Functions]
\label{theorem-canonical-combinations}
Consider a set of convex functions $f_1,\ldots,f_n$ mapping $\+x \to \mathbb{R}$, and $\alpha_1,\ldots,\alpha_n$ be a set of non-negative scalars, then:
\[ g(\+x) = \sum_{i=1}^n a_i f_i = a_1 f_1 + a_2 f_2 + \ldots + a_n f_n\]
is also convex. Furthermore, if any $f_i$ is strictly convex, then $g(\+x)$ is strictly convex.
\end{theorem}

The concavity of the joint acquisition function indicates the existence of its global optimum. Next, we will demonstrate the existence of Pareto optimal in the MOO setting.


\begin{theorem}[Sufficient Condition for Pareto Optimality \cite{Marler2004SurveyMOO}]
\label{theorem-Sufficient-Condition}

Let $F \in Z$, $x^* \in X$, and $F^* = F(x^*)$. Let a scalar global criterion $F_g(F) : Z \rightarrow \mathbb{R}$ be differentiable with $\nabla_{F} F_g(F) > 0 \ \forall F \in Z$. Assume $F_g(F^*) = \min\{F_g(F) < F \in Z\}$. Then, $x^*$ is Pareto optimal.

\end{theorem}

Define that \(Z = \{\+S, \+U\}\), \( x^* \in X \), the joint acquisition function is the scalar global criterion \( \+J : Z \rightarrow \mathbb{R} \), we need to prove that \( \+J \) increases monotonically with respect to \( \+S, \+U \), which means that \( \nabla_{F} \+J > 0 \) for all \( F\in Z\). Since \( \partial \+J / \partial \+S = \+U \), the uncertainty score \( \+U \) is based on entropy so \( \partial \+J / \partial \+S > 0 \). Similarly, since \( \partial \+J / \partial \+U = \+S \), the similarity score \( \+S \) is designed based on the triplet loss with cosine similarity as a distance function, so \( \partial \+J / \partial \+U > 0 \) intuitively. Therefore, we have $\nabla_{F} \+J > 0, \text{for all } F\in Z$, for our proposed joint acquisition function. Following Theorem \ref{theorem-Sufficient-Condition}, we have the optimal solution of a global function \( \+J \) is sufficient for achieving the Pareto optimality of separate objectives if \( \+J \) increases monotonically with respect to each objective function.

Given the existence of the optimum of joint acquisition function and the Pareto optimal of separate objectives, we further demonstrate the design of AL can converge to the optimal point with the increase of queried samples. Referring to Equation (\ref{joint-score-vec-eq}), the similarity scores serve as a filter to keep uncertainty scores for those unlabelled samples that are close to the target population. It essentially reduces the problem into a classical AL setting, where no distribution mismatch exists (data samples from different distributions are eliminated). Raj et al.\cite{raj2021convergence} show that the uncertainty sampling in classic AL converges to the optimal predictor for binary classification and provide the provide a non-asymptotic rate of convergence of order $O(1/n)$, where n is the number of iterations of the AL.

\section{Case Study} \label{section-experiments}

This section presents the motivation, core design choices, as well as a comprehensive overview of the experiments conducted as a case study to verify the effectiveness of the proposed ADs framework in industrial tasks. 

\subsection{Experimental Setup}

ML applications in industrial tasks are generally hindered by challenges, including data scarcity, limited annotation budgets, and lack of prior knowledge with regard to data distribution and informativeness, which leads to the shared data is not aligned with the target distribution and uninformative. Data sharing is commonly purported as a solution, but it only addresses the problems of data scarcity and, to an extent, annotation budget. The issues of distribution mismatch and uninformative of the shared data remain prevalent, thereby delaying the performance of the downstream task. It follows that the purpose of any framework attempting to solve these is to further satisfy that (1) the distribution of the selected data matches the target distribution, i.e., the distribution mismatch of shared data is minimized, and (2) the selected data contributes valuable information towards the model dealing with the downstream task.

\begin{figure}[!h]
    \centering
    \includegraphics[width=3.4in]{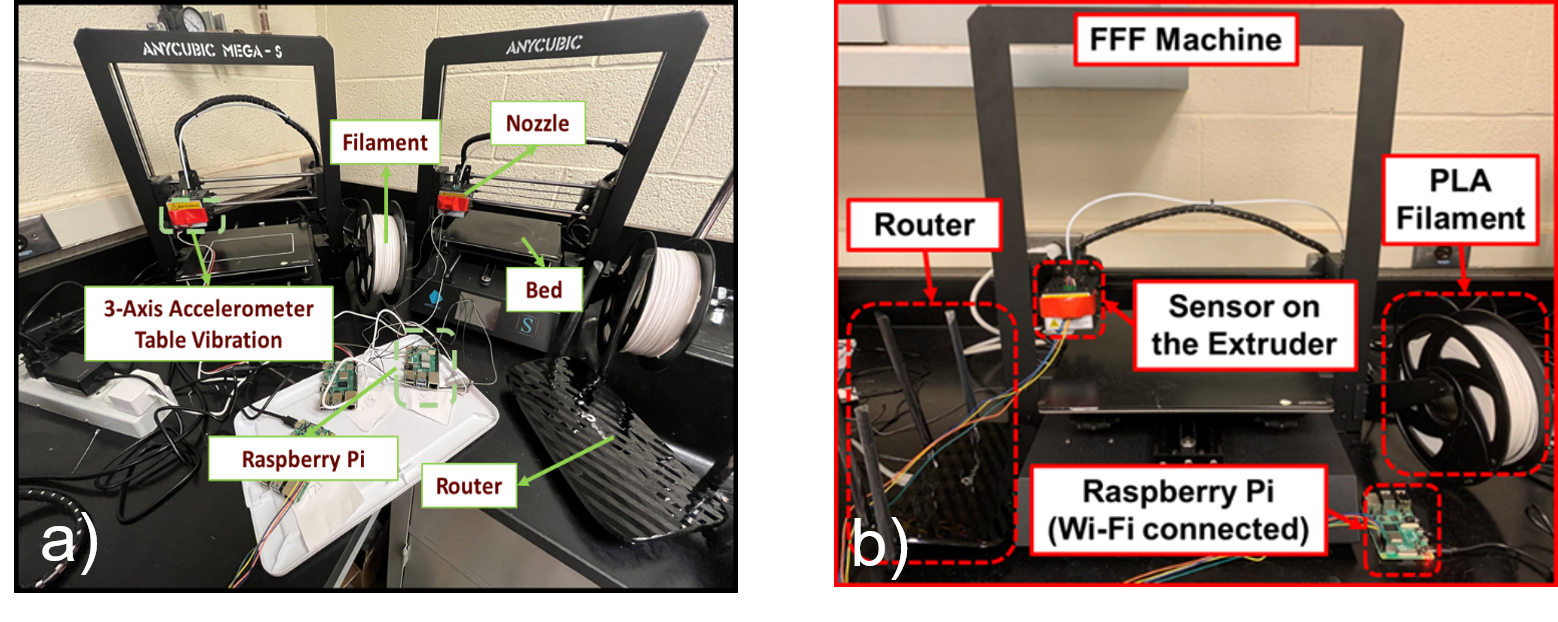}
    \caption{Experimental platform using a Fused Filament Fabrication (FFF) machine (a) two small printers of the same make and model, and (b) one large printer from a separate manufacturer \cite{shi2023knowledge}.}
    \label{fig-additive-machine}
\end{figure}

The problem scenario for the case study is carefully designed to simulate these issues in the real world. Data is collected for the same additive manufacturing process from three machines with the ultimate goal of enabling data-sharing to enhance the downstream task of anomaly detection in the manufactured object. The particular anomaly to detect is a manufacturing fault, like the one shown in Figure \ref{fig-am-printed-product}. Research has shown that the monitoring data can reflect this anomaly \cite{li2021augmented}. Of the three machines, two are smaller in size and of the same make and model, and the third is a larger model from a separate manufacturer, as evident from Figure \ref{fig-additive-machine}. In this section, the two similar small machines are termed $S = \{S_1, S_2\}$, and the dissimilar large machine is termed $L$. 

The purpose of having two similar and one dissimilar machine is to explicitly introduce the distribution mismatch in a typical data-sharing problem. Specifically, the case study considers the scenario where all three machines are tasked to manufacture the same object, a cube with dimensions $2\times2\times2 \text{cm}^3$. It is natural to treat the monitoring data obtained from similar machines following a similar distribution compared to that from the dissimilar machine owing to the variety in build design, software, specifications, and other manufacturing differences. Therefore, distribution mismatch needs to be considered when sharing the monitoring data from these three machines with each other. As discussed earlier, this is the issue that the contrastive model in AD addresses. Furthermore, given that the data is scarce and annotations are costly in terms of manual time and effort, it becomes critical to have only the most informative samples annotated for the downstream task. The uncertainty sampling model of ADs is well-suited for this task. The model requires a classifier to identify normal and abnormal data, which, in this work, is based on a Convolutional Neural Network (CNN) architecture with cross-entropy loss and Adam optimization\cite{KingBa15}.

\subsection{Data Preprocessing and Augmentation} \label{subsection-wtk}

In the ADs framework, the initial phase involves preprocessing the data and training the augmentation model. This model is specifically designed to generate positive pairs for time sequence data, serving as the training input for the CL model. The entirety of the data is min-max normalized to ensure consistent scaling throughout. An initial subset of the complete data (e.g., 20\%) is then sampled for manual annotation. Rather than use random sampling, ADs employs Latin Hypercube Sampling (LHS) \cite{eglajs1977new, mckay2000comparison} to ensure that the sampled data is uniformly distributed within the parameter space and therefore representative of the real variability of the distribution. Following this, the Winner-Take-All autoencoder of \cite{makhzani2015winner} is trained in an unsupervised fashion to learn shift-invariant and sparse, high-dimensional latent embeddings of the normalized data. Once trained, transformed augmentations of the input data are generated by passing the input through the encoder to generate the latent input embedding, manipulating it in the latent space, and finally reconstructing the input by decoding the modified latent embedding. The reconstruction result serves as the transformed augmentation. ADs defines two augmentation transformations in this way: 

\textit{Additive Gaussian Noise:} The first augmentation $\+y_{a_1}$ of the $n$-dimensional latent space output embedding $\+y_e$ is obtained by adding a Gaussian noise vector $\+n$ to $\+y_e$: 
\begin{equation}\label{augmentation-1} 
\+y_{a_1} = \+y_e + \+n 
\end{equation}
\[ \+n = \frac{r \sigma_e}{5} \times \+M \]

Where $r \sim \mathcal{N}(0,1)$, $\sigma_e$ indicates the standard deviation of the latent space output, and $\+M$ is a 2D binary mask array based on $\+y_e$:
\[
\+M = \left\{ \begin{array}{lcr}
    1 & \text{if} & y > \sigma_e \\
    0 & \text{if} & y \leq \sigma_e
\end{array}, \;\forall\;y \in \+y_e \right. 
\]

Shortly, $M$ enforces that noise is only added to those components of the output embedding $\+y_e$ that lie outside the range of one standard deviation of the vector. This helps ensure that the generated augmentation exploits the variability of the embedding as much as possible by adding noise only to the components that deviate significantly from the mean. Thus, the generated augmentation retains components most closely clustered around the mean, thereby preserving the most characteristic features. This is desirable as the augmentations are meant to be used as positive examples while training the contrastive network. Consequently, they should not deviate too much from the actual sample (i.e., the anchor) to mitigate the risk of generating an out-of-distribution (OOD) augmentation.

\textit{Embedding Deviation Thresholding:} The second augmentation $\+y_{a_2}$ is generated by multiplying $\+y_e$ with its absolute value $\lvert\,\+y_e\rvert$ to produce a resultant vector $\+y_e^\prime$. Each component of squared components $\+y_e^\prime$ is then compared with $\sigma_e$, such that all components greater than $\sigma_e$ are set to 1 and the rest to 0. Mathematically:
\[ \+y_e^\prime = \+y_e \odot |\,\+y_e| \]
\begin{equation}  \label{augmentation-2}
\+y_{a_2} = \left\{ \begin{array}{lcr}
    1 & \text{if} & y > \sigma_e \\
    0 & \text{if} & y \leq \sigma_e
\end{array}, \;\forall\;y \in \+y_e^\prime \right. 
\end{equation}

This augmentation is similar to the mask in the first one, with the main differences being that no noise is added, and the threshold is applied to the \textit{squared} components of the output embedding, thereby exaggerating the distance, or lack thereof, of each component from the mean.

\begin{figure}[!h]
    \centering
    \includegraphics[width=2.45in]{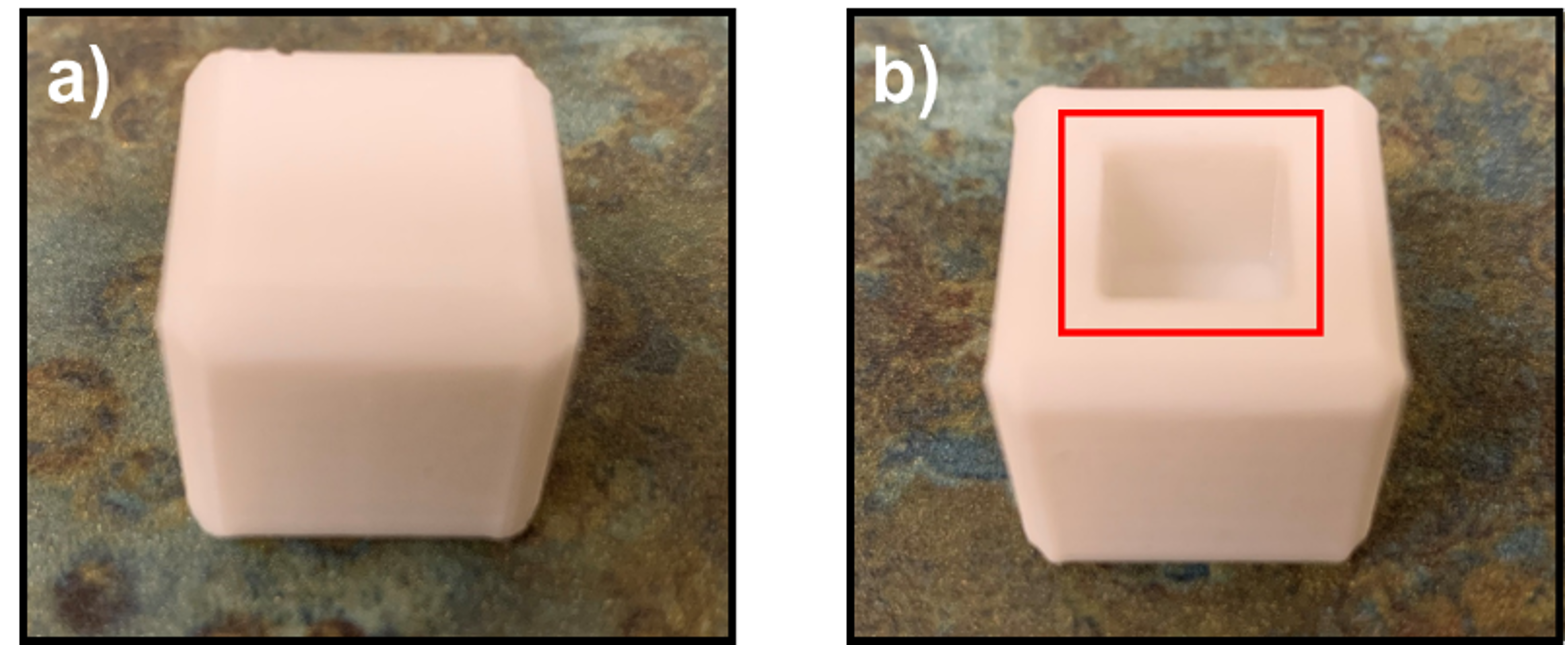}
    \caption{The printed product, a \(2 \times 2 \times 2 \text{ cm}^3\) cube. The square void indicated in the red bounding box is the anomaly that the downstream model is required to detect.}
    \label{fig-am-printed-product}
\end{figure}

In order to evaluate the effectiveness of ADs as well as perform a comparative analysis with various other settings and methods, data from the additive manufacturing machines were acquired as they printed a cube with a small square void as the anomaly on one of the faces as shown in Figure \ref{fig-am-printed-product}. It is important to understand the data handling procedure to ensure the quality of the evaluation. To that end, a few terms are now introduced. \textit{Source identification} is defined as the task of identifying whether a sample belongs to $S$ or $L$, and \textit{Classification} as assigning it the normal or abnormal class. For ease of reference, data may be considered notorious, i.e., it is unknown which machine produced them As the first step in data pre-processing for this case study, the actual identity and class labels of all position recordings are isolated and stored separately to compute evaluation metrics. As such, the incoming data is both anonymous and unclassified (unlabeled). The data is then combined into a single, large dataset that is hereafter referred to as the initial dataset. Additionally, to make the task more challenging, the ratio of the data $L:S$ is kept large so that the majority of the combined dataset is populated by data from the dissimilar machine. Finally, since the application of a supervised ML approach requires some annotated data, a small initial subset of the data is sampled via LHS and annotated so that almost all of the data remains unlabeled. 

These \textit{initial} annotations consist of both source identification ($S$ or $L$) as input for a CL model and data classification (normal or abnormal). Note that source identification is not related to the downstream task of anomaly detection -- it is only included to allow training the CL model in the ADs architecture. All successive annotations in the AL cycle consist purely of normal/abnormal classification so that the classifier can be retrained on the high-entropy data.

\subsection{Experiment Settings}

\begin{table*}[!h]
\centering
\caption{Experiment settings by using the shared data from the three additive manufacturing machines.
}
    \vspace{-1mm}
    \begin{tabular}{cccccc}
        \toprule
        &$\mathcal{D}_{inital}$ & $\mathcal{D}_{training}$ & $\mathcal{D}_{testing}$   \\
        \midrule
        \bottomrule \\[-1.9ex]
        Supervised & N/A & $ \{S_1,S_2\}^\text{100\% train}$ & $ S_1^\text{test},S_2^\text{test} $  \\
        Random Pick S & $20\% S_1, 20\%S_2, 20\%L$ & $ \{S_1,S_2\}^\text{random pick}$ & $ S_1^\text{test},S_2^\text{test} $  \\
        Random Pick S+L & $20\% S_1, 20\%S_2, 20\%L$ & $ \{S_1,S_2,L\}^\text{random pick}$ & $ S_1^\text{test},S_2^\text{test} $  \\
        ADs Without CL & $20\% S_1, 20\%S_2, 20\%L$  & $ \{S_1,S_2,L\}^\text{queried}$ & $S_1^\text{test},S_2^\text{test}$ \\
        \textbf{ADs} &  \text{$20\% S_1, 20\%S_2, 20\%L$}  & \text{$ \{S_1,S_2\}^\text{queried} $ }  & \text{ $ S_1^\text{test},S_2^\text{test} $ } \\
        \bottomrule
        
    \end{tabular}
    \label{table-Experiment-settings}
\end{table*}

At this point, the data is ready for input to ADs. Comparative experiments among the proposed ADs and other benchmark methods for anomaly detection are conducted for more conclusive and insightful results. To that end, anomaly detection is performed in five different settings that are now discussed in order. The experiment setting is shown in Table \ref{table-Experiment-settings}. Note that, in all settings: (1) In-text variables should be considered specific to the setting they are in. (2) Testing dataset for all settings was kept the same for a fair comparison. (3) Per typical training regulations, the training, testing, and validation sets $\mathcal{D}_\text{train}, \mathcal{D}_\text{test}, \mathcal{D}_\text{val}$ are pairwise disjoint.
 
\textbf{\textit{Supervised:}} This is the usual setting for most deep-learning approaches dealing with classification. It assumes that labels are available for all samples, which is the opposite of our scenario as it implies that data collection and annotations are not a problem, so data-sharing is not needed. As such, its performance on the downstream task is considered the baseline. 

\textbf{\textit{Random Pick S:}} In this setting, only the data from similar machines is made available to the model, so $\mathcal{D} = \{S_1, S_2\}$. By removing the dissimilar data from the dataset, this setting allows observing the effect of the uncertainty score $\+U$ from ADs in isolation, thus evaluating its effectiveness. 

\textbf{\textit{Random Pick S+L:}} This is similar to the \textit{Random Pick S} setting, with the main difference being that the data available to the model is provided by all three machines, so $\mathcal{D} = \{S_1,S_2,L\}$. As such, this setting better allows studying the usefulness of both the similarity score $\+S$ and uncertainty score $\+U$. 20\% of the entire data alongside 400 randomly picked samples from the remaining data forms the subset $\mathcal{D}^\text{sub} \subset \mathcal{D}$. Evidently, only the training dataset changes.

\textbf{\textit{ADs Without CL:}} This setting strips away the contrastive model from the ADs framework so that instead of a joint score $\+J$ comprised of a similarity score $\+S$ and uncertainty score $\+U$, each sample is only assigned $\+U$ to generate the query set for the next cycle. Thus, it represents a purely AL-based approach to data-sharing. In this setting, data is sourced from all three machines, so $\mathcal{D} = \{S_1,S_2,L\}$. Like in the \textit{Random Pick S+L} settings, a subset of the data is formed by sampling 20\% of this data, but no random samples are added. Instead, the 400 new samples are selected via AL. ADs is set to terminate after 5 cycles with 80 samples per cycle to ultimately select these 400 samples, and after adding them to the sampled 20\%, the final subset of data $\mathcal{D}^\text{sub} \subset \mathcal{D}$ for training and validation is ready. Evidently, the only difference is how the 400 samples are selected. 

\textbf{\textit{ADs:}} This setting is exactly the same as \textit{ADs Without CL}, except for keeping the contrastive model in the ADs framework so that each sample is selected based on the joint score $\+J$, which is in turn computed via an integration of the similarity score $\+S$ and uncertainty score $\+U$. Consequently, this setting minimizes the number of dissimilar data $L$ introduced into the anomaly detector's training while still prioritizing data informativeness and thereby boosting model accuracy. Evidently, ADs represents AL as a multi-objective optimization problem, resulting in selected samples that are not only high-entropy samples but also highly likely to originate from similar machines $S$. 

\textbf{\textit{Random Pick}}: It denotes that training data (20\% of the full data) is still specifically extracted, but the additional samples are selected at random. This decision is logically informed: one might imagine that owing to the randomized nature of the data, the anomaly detector model would form different decision boundaries when re-training with a new data split. Therefore, by running multiple instances of a model trained as such, the average performance of these models leads to a more accurate estimate of the true performance of the model.

\subsection{Results}

\begin{figure}[!h]
\begin{centering}
\subfloat[]{\includegraphics[scale=0.25]{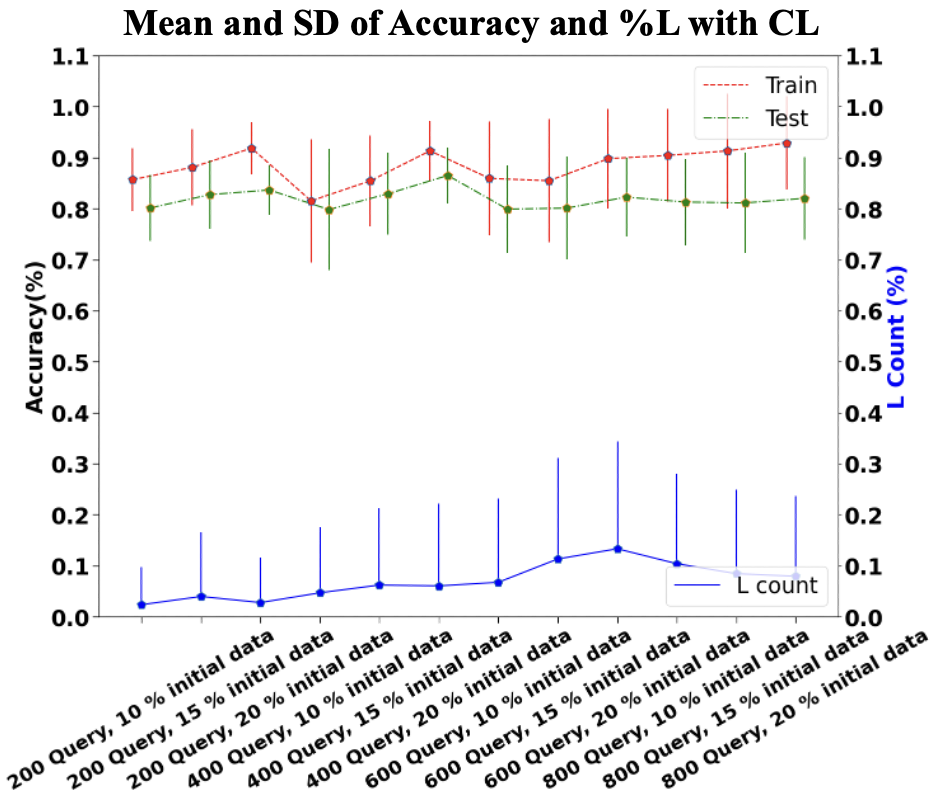}

}\par\end{centering}

\begin{centering}

\subfloat[]{\includegraphics[scale=0.25]{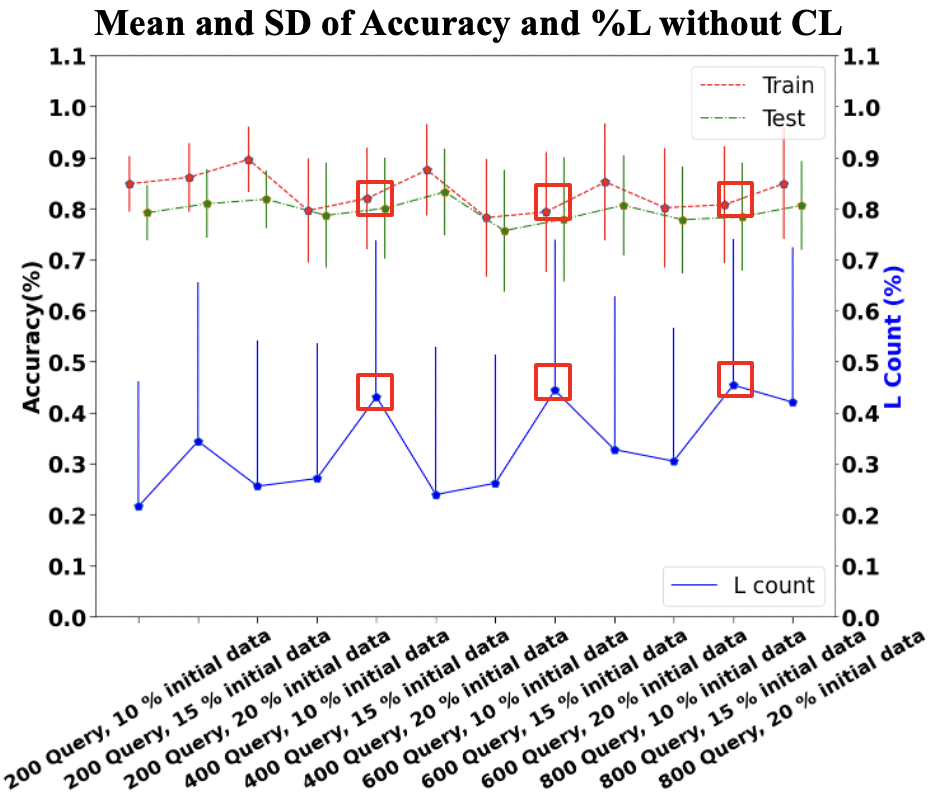}

}
\par\end{centering}

\caption{The ADs Framework efficiently queried high-quality data and also did not require a full dataset to achieve promising performance. (a) The usage of similarity scores computed from the CL network corresponds to lower \%L, which in turn provides higher accuracy as expected. (b) Without CL and similarity scores, we see much higher \%L values, resulting in worse accuracy and greater variability of the performance.}
\label{fig-result-graph}
\end{figure}

The corresponding results, including the anomaly detector's accuracy, F1 score, and the percentage of samples that are not from the target distribution (\textit{\%L}), are tabulated in Table \ref{table-case-study-results}. Ideally, the classification accuracy would be 100\% (i.e., all selected samples are correctly classified as normal or abnormal), and the percentage of negative samples in the selected unlabeled samples would be 0\% (i.e., there are no negative samples in the selected samples). Furthermore, the effect of \textit{\%L} on the model accuracy is visualized in Figure \ref{fig-result-graph} for different choices of queried data and the percentage of complete data used to create the training subset (\textit{Q. Data} and \textit{\%Data} in Table \ref{table-case-study-results}). Note that \textit{\%L} is only applicable to \textit{Random Pick S+L, ADs without CL and ADs} settings, which involve the dissimilar data from the larger machine $L$ and also present a chance of incorrectly adding its samples into the training pipeline. This is unlike the settings \textit{Supervised, Random Pick S+L}, which only deal with similar machine data.

\begin{table}[!h]
\centering

\caption{Results of the case study by using the shared data from the three additive manufacturing machines.}
    \vspace{-1mm}
    \begin{tabular}{cccccc}
        \toprule
        & \%Data & Q. Data & Accuracy & F1 Score & \%L \\
        \midrule
        \bottomrule \\[-1.9ex]
        Supervised & 100 & n/a & 0.9437 & 0.9449 & n/a \\
        Random Pick S & 20 & 400 & 0.8957 & 0.8980 & n/a \\
        Random Pick S+L & 20 & 400 & 0.8805 & 0.8805 & 45.25 \\
        ADs Without CL & 20 & 400 & 0.9510 & 0.9507 & 12.5 \\
        \textbf{ADs} & \textbf{20} & \textbf{400} & \textbf{0.9578} & \textbf{0.9563} & \textbf{0}\\
        \bottomrule
    \end{tabular}
    \label{table-case-study-results}
\end{table}


As evident from the results in Table \ref{table-case-study-results}, even with just 20\% of the complete data and 400 additional samples selected using our novel query strategy, semi-supervised ADs outperform the supervised baseline, both with and without CL (i.e., whether the similarity score was involved in the joint score computation). This is evidenced specifically by the increase in accuracy (ADs: $+1.4\%$, ADs Without CL: $+0.73\%$) and F1 scores (ADs: $+1.14\%$, ADs Without CL: $+0.58\%$). Additionally, the results from Figure \ref{fig-result-graph} confirm that enabling CL to generate similarity scores effectively allows the framework to distinguish similar and dissimilar data, evidenced by the significantly lower \textit{\%L} scores in the setting with CL enabled. Additionally, lower \textit{\%L} values consistently correspond to higher accuracy in the downstream anomaly detection task despite variations in the size of the training data. On the contrary, the setting with CL disabled picks large amounts of dissimilar data, which corresponds to generally worse performance as well as a greater degree of result variability.

\subsection{Ablation Study}

\begin{figure*}[ht]
\begin{centering}
\subfloat[]{\includegraphics[scale=0.32]{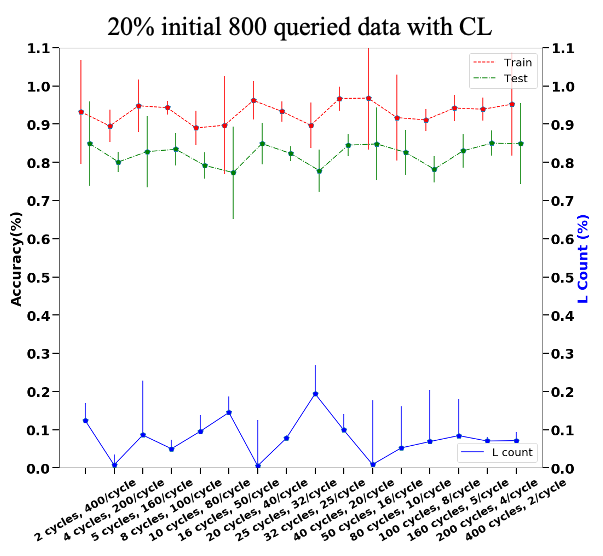}

}\subfloat[]{\includegraphics[scale=0.32]{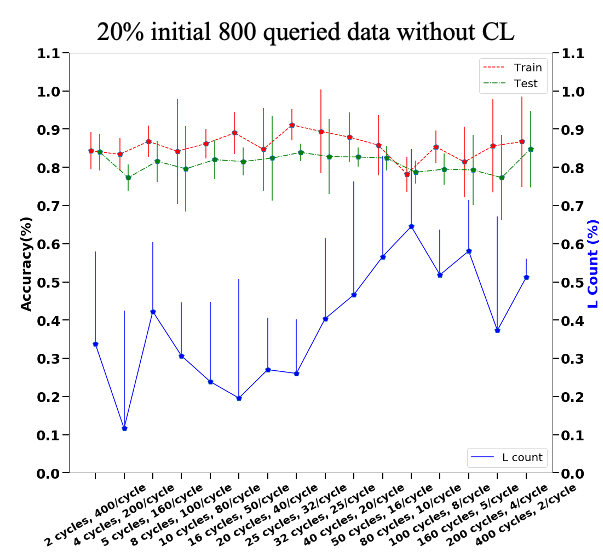}

}
\par\end{centering}
\begin{centering}
\subfloat[]{\includegraphics[scale=0.32]{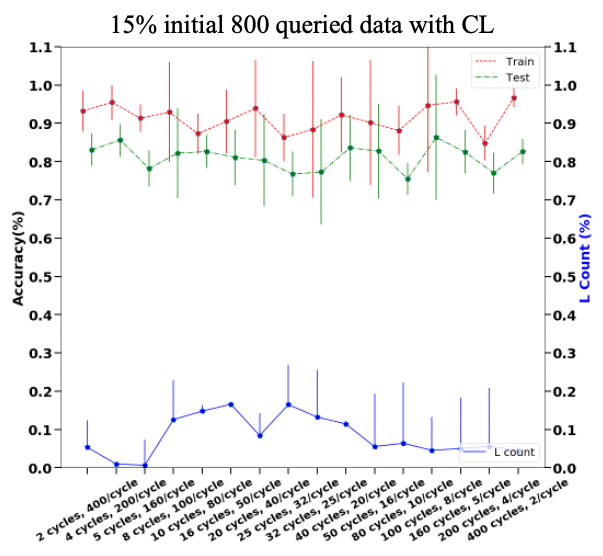}

}\subfloat[]{\includegraphics[scale=0.32]{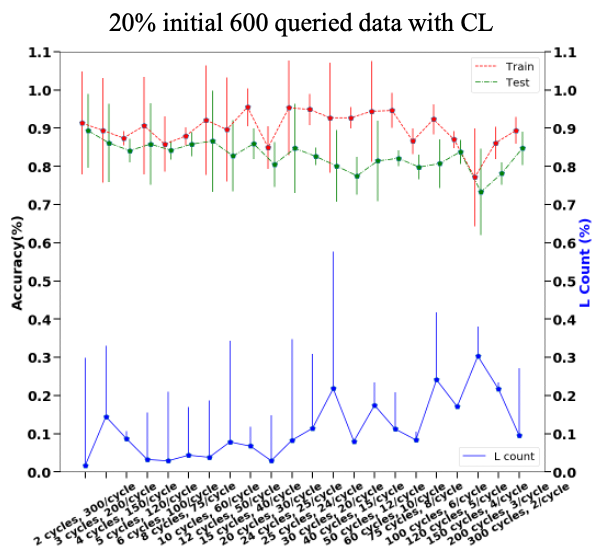}

}
\par\end{centering}
\caption{Results of the ablation experiments. (a) and (b) evaluate the performance with and without CL, (c) evaluate the performance with varying numbers of initially sampled data, and (d) evaluate the performance with varying total numbers of selected samples.}
\label{fig-ablations}
\end{figure*}


In order to further reinforce the results of the main experiments, ablation experiments were performed. The goal of these is to (1) establish that the reduction in selection of dissimilar samples \textit{\%L} is not related to the number of selected samples and primarily concerned with the usage of CL, (2) study the relationship between model performance and the number of selected samples per cycle of ADs for a fixed budget of total samples, and (3) study the effect of increasing and decreasing the amount of queried data in each cycle.To accomplish the first two objectives, the number of total queried samples was fixed to 800 in addition to 20\% initially sampled data, and the only two variables are whether CL is enabled or not and the number of samples selected per cycle of ADs. There are 16 experiments conducted, and the total queried budget is 800. The samples were evenly queried by each cycle and the number of cycles selected in set \{2,4,5,8,10,16,20,25,32,40,50,80,100,160,200,400\} respectively.

The results show a consistently and significantly smaller \%L for the model with CL as compared to without CL. The fewer the number of samples queried per cycle, the more likely the model has higher values of \%L without CL. With CL, the result remains fairly similar regardless of the number of samples queried per cycle. This confirms the hypothesis that CL is the primary factor in reducing the number of dissimilar samples in the final query set, for these ablation experiments are visualized in Figures \ref{fig-ablations}(a) and \ref{fig-ablations}(b). For the third goal, CL is kept enabled, and the number of queried samples is 600 and 800, with 20\% initial data in both cases. The model performance is compared over varying samples per cycle for each scenario. While the performance does improve for the case with a larger budget, the difference is considerably small, which is 200 queried samples difference between the two settings. The results of this experiment are visualized in Figures \ref{fig-ablations}(a) and \ref{fig-ablations}(c). We also experimented with different numbers of initial data by keeping the CL enabled and the same number of queried samples. The model performance with 15\% and 20\% initial data are similar. The results of this experiment are visualized in Figures \ref{fig-ablations}(a) and \ref{fig-ablations}(d).

\section{conclusion}
Data-sharing among machines has great potential to contribute to the wide application of ML methods in manufacturing systems, which addresses the challenges of data scarcity. However, typical data-sharing approaches do not consider the quality of the shared samples, nor the inherent distribution mismatch of the data thus acquired. In this paper, a semi-supervised Active Data-sharing (AD) framework is proposed to address these problems. ADs selects high-quality data that are both informative to the downstream ML task and appear to follow the target distribution. ADs views the problem as a multi-objective optimization problem and employs a novel joint acquisition function to query the Pareto optimal point that satisfies both objectives simultaneously. The joint score itself is computed based on a combination of two individual scores, namely the uncertainty and similarity scores, that are separately obtained via entropy and CL techniques. 

Systematic experiments are conducted using ADs to evaluate its effectiveness. The experiment involves real-world in-situ monitoring data from the same additive manufacturing process using data-sharing between three machines, two of which are similar, i.e., the same model, and one large machine from a different manufacturer is considered dissimilar to purposely introduce distribution mismatch. Results show that with only a fraction of initially annotated data and a few cycles of ADs to extend the annotated dataset, the anomaly detection model outperforms the baseline anomaly detector trained on the fully annotated dataset. Furthermore, the excel performance is achieved in a distribution-aware manner, i.e., without querying any of the samples from different distributions to use for training, thereby effectively addressing the distribution mismatch problem. Further ablation studies confirm that the design philosophy of ADs with the combination of AL and CL is indeed the factor in improving performance. The usage of the similarity score directly reduces the mismatched data in the selection, and smaller amounts of mismatched data further improve the accuracy of the anomaly detection task. 

In light of these extensive experiments and their results, it is concluded that ADs effectively solve the problems of data quality and distribution mismatch usually prevalent in existing data-sharing approaches and that this work establishes a new baseline for further improvements to data-sharing in the industrial domain.

\appendix
\section{Appendix}\label{Appendix}

\subsection{Concavity of Entropy} \label{proof-entropy-concavity}
Assuming a C-way classification problem, where the model's output probabilities are defined in terms of the model parameters as $P_\theta$ and the predicted class is $\hat y_i$, the Shannon entropy is defined as:
\begin{flalign*}
H(x) &= \sum_i^C P_\theta(\hat y_i\;|\;x) \log_b\left(\frac{1}{P_\theta(\hat y_i\;|\;x)}\right)  \\
&= -\sum_i^C P_\theta(\hat y_i\;|\;x) \log_b(P_\theta(\hat y_i\;|\;x)) 
\end{flalign*}

Where $i$ represents the $i^\text{th}$ class and $C$ represents the total number of classes. In binary classification problems (as in the case presented in this paper), the summation symbol can be discarded:
\[ H(x) = - P_\theta(\hat y\;|\;x) \log_b(P_\theta(\hat y\;|\;x)) \]

There are several methods to prove the concavity of Shannon entropy. For readability, let $P_\theta(\hat y\;|\;x) = p_x$. Given that for a function to be concave, the second derivative of the function with respect to all its variables must be non-positive over the entire domain of the function:
\begin{align}
\label{entropy-derivative}
    H^\prime(x) & = \frac{d}{dp_x}\left(-p_x\log_b p_x\right) \nonumber \\
    & = -\left(\log_b p_x + \frac{p_x}{p_x\ln b}\right) \nonumber \\
    & = -\left(\log_b p_x + \frac{1}{\ln b} \right) \nonumber \\
    H^{\prime\prime}(x) & = \frac{d}{dp_x}\left(-\log_b p_x - \frac{1}{\ln b} \right) \nonumber \\
    & = -\frac{1}{p_x \ln b}
\end{align}

Here, the probability mass $p_x \in [0,\,1]$ and therefore non-negative by definition, while $\ln(b)$ is a non-negative constant for $b > 1$ as is the case with logarithmic bases. Therefore, the expression in Equation \ref{entropy-derivative} always evaluates to a non-positive value, which proves that entropy is a concave function.

\bibliographystyle{ieeetr}
\bibliography{IEEEabrv,bibliography}

\end{document}